\documentclass[letterpaper]{article} 
\usepackage[draft]{aaai25}  
\usepackage{times}  
\usepackage{helvet}  
\usepackage{courier}  
\usepackage[hyphens]{url}  
\usepackage{graphicx} 
\usepackage{natbib}  
\usepackage{caption} 
\usepackage{algorithm}
\usepackage{algorithmic}
\usepackage{amsmath}
\usepackage{amsfonts}
\usepackage{booktabs}
\usepackage{newfloat}
\usepackage{listings}
\DeclareCaptionStyle{ruled}{labelfont=normalfont,labelsep=colon,strut=off} 
\floatstyle{ruled}
\newfloat{listing}{tb}{lst}{}
\floatname{listing}{Listing}
\title{Scaling Combinatorial Optimization Neural Improvement Heuristics with Online Search and Adaptation}
\author{
    Federico Julian Camerota Verdù \textsuperscript{\rm 1},
    Lorenzo Castelli\textsuperscript{\rm 2},
    Luca Bortolussi\textsuperscript{\rm 1}
}
\affiliations{
    \textsuperscript{\rm 1}Dipartimento di Matematica, Informatica e Geoscienze, Università degli Studi di Trieste\\
    \textsuperscript{\rm 2}Dipartimento di Ingegneria e Architettura, Università degli Studi di Trieste\\

    federicojulian.camerotaverdu@phd.units.it
}

\usepackage{bibentry}

\begin{document}

\maketitle

\begin{abstract}

We introduce Limited Rollout Beam Search (LRBS), a beam search strategy for deep reinforcement learning (DRL) based combinatorial optimization improvement heuristics. 
Utilizing pre-trained models on the Euclidean Traveling Salesperson Problem, LRBS significantly enhances both in-distribution performance and generalization to larger problem instances, achieving optimality gaps that outperform existing improvement heuristics and narrowing the gap with state-of-the-art constructive methods.
We also extend our analysis to two pickup and delivery TSP variants to validate our results.
Finally, we employ our search strategy for offline and online adaptation of the pre-trained improvement policy, leading to improved search performance and surpassing recent adaptive methods for constructive heuristics. Our source code is available at \url{anonymous-url}.
\end{abstract}

%

\section{Introduction}
Combinatorial Optimization (CO) problems can be found in several domains ranging from air traffic scheduling~\cite{bertsimas2011integer} and supply chain optimization~\cite{singh2022combinatorial} to circuit board design~\cite{barahona1988application} and phylogenetics~\cite{catanzaro2012balanced}.
Although general-purpose solvers exist and most CO problems are easy to formulate, in many applications of interest getting to the exact optimal solution is NP-hard and said solvers are extremely inefficient or even impractical due to the computational time required to reach optimality~\cite{TOTH2000222, COLORNI19961}.
Specialized solvers and heuristics have been developed over the years for different applications. 
However, the latter are often greedy algorithms based on hand-crafted techniques that require vast domain knowledge, thus they cannot be used on different problems and may get stuck on poor local optima~\cite{applegate2003chained,helsgaun2009general, gasparin2023evolution}.

CO problems have gained attention in the last few years within the deep learning community where neural networks are used to design heuristics that can overcome the limitations of traditional solvers~\cite{lombardi2018boosting, bengio2021machine}.
In particular, an extensive literature has been developed on methods to tackle the travelling salesperson problem (TSP) due to its relevance and particular structure that allows to easily handle constraints with neural heuristics.
Deep learning approaches for CO problems can be divided into constructive and improvement methods. 
The former follows a step-by-step paradigm to generate a solution starting from an empty one and sequentially assigning decision variables~\cite{vinyals2015pointer,nazari2018reinforcement,kool2018attention}. 
Instead, improvement approaches iteratively improve a given initial solution using an operator to turn a solution into a different one~\cite{zhang2020learning,d2020learning,wu2021learning,HOTTUNG2022103786}.
Moreover, deep learning solvers can be classified based on their learning strategies: supervised learning~\cite{khalil2017learning,joshi2019efficient, hottung2020learning,  li2021learning, xin2021neurolkh, sun2023difusco} and deep reinforcement learning (DRL)~\cite{bello2017neural,khalil2017learning,deudon2018learning,kool2018attention, ma2019combinatorial,barrett2020exploratory,kwon2020pomo,d2020learning,kim2021learning,ma2021learning,qiu2022dimes,ye2023deepaco,grinsztajn2023winner, ma2023neuopt}.

Many recent advancements in neural solvers for CO primarily lie within the constructive framework.
This approach eliminates the necessity for manually crafted components, thereby providing an ideal means to address problems without requiring specific domain knowledge~\cite{lombardi2018boosting}.
However, improvement heuristics can be easier to apply when complex constraints need to be satisfied and may yield better performance than constructive alternatives when the problem structure is difficult to represent~\cite{zhang2020learning} or when known improvement operators with good properties exist~\cite{bordewich2008consistency}.
Still, generalization, i.e., scaling from training sets with small problems to large instances while retaining good performance, is an open issue when using DRL neural heuristics in CO, particularly for the TSP~\cite{joshi2021learning}.

\paragraph{Contributions.}
While generalization has been studied for constructive methods~\cite{hottung2021efficient,oren2021solo,choo2022simulation, son2023meta, jiang2023ensemblebased, li2023distribution}, to the best of our knowledge no prior work has been done on improvement heuristics.
In this paper, we focus on improvement heuristics for the TSP based on DRL policies and propose an inference-time beam search approach, Limited Rollout Beam Search (LRBS), which allows tackling problems $10$ times larger than those seen at training time.
Using pre-trained models, on instances of the same size as those used for training, our search scheme achieves state-of-the-art results among similar improvement methods and shows comparable performance to constructive heuristics.
Generalization to instances up to $10$ times larger than those seen while training is improved considerably with respect to sampling from the original policy, mitigating the gap with constructive solvers.
Moreover, our approach allows the integration of online adaptation within the search to overcome the limitations posed by the pre-trained model in large-scale generalization. 
We also investigate the effectiveness of LRBS as an exploration strategy in fine-tuning the pre-trained models on a limited dataset of instances of the same size as those in the test set.
In this setting, our experiments show competitive performances with constructive approaches that use online instance-based adaptation.  
Finally, we validate LRBS on two pickup and delivery TSP variants and show the advantage of our search approach with respect to more specialized problem-specific solutions. 
In conclusion, our analysis indicates that solvers utilizing improvement heuristics and a robust exploration approach may offer a viable alternative to adaptive constructive methods, displaying enhanced scalability for larger problem instances in terms of computational times.

\section{Preliminary and Related Work}
This section begins by introducing the fundamental concepts of the DRL framework for learning improvement heuristics. 
Subsequently, we discuss relevant search and adaptive methods in the neural CO literature.

\subsection{Improving TSP Solutions with DRL}
\label{subsec:improv}
A TSP instance is defined by a graph $G=( V,\ E)$ and the objective is to find a tour $\delta$, i.e. a sequence of nodes $x_i \in V$, such that each node is visited only once, the tour starts and finishes in the same node and minimizes the tour length $$L(\delta)= w_{\delta_N,\delta_1} + \sum_{i=1}^{N-1} w_{\delta_i, \delta_{i+1}},$$ where $N = |V|$, $w_{ij} \in \mathbb{R}^+$ and $(i,\ j) \in E$ are edges in the graph.
In this work, we consider instances of the Euclidean TSP~\cite{arora1996polynomial} where $w_{ij} = \|x_i - x_j\|$. 

To solve TSP instances in the improvement framework we start from a given randomly generated initial solution $\delta^0$ and use a policy $\pi_\theta$, parametrized by learnable weights $\theta$, to sequentially improve $\delta^0$. 
The policy selects actions in the neighbourhood defined by an operator $g$ that, given a solution and an action, returns another solution to the problem. 
We formulate the DRL framework as follows.

\paragraph{State.} The state is given by the current solution $\delta^T$ and the best solution found so far $\delta^{*T}=\text{argmin}_{i\leq T} L(\delta^i)$.

\begin{figure}[ht]
\centering
\centerline{\includegraphics[width=0.7\columnwidth]{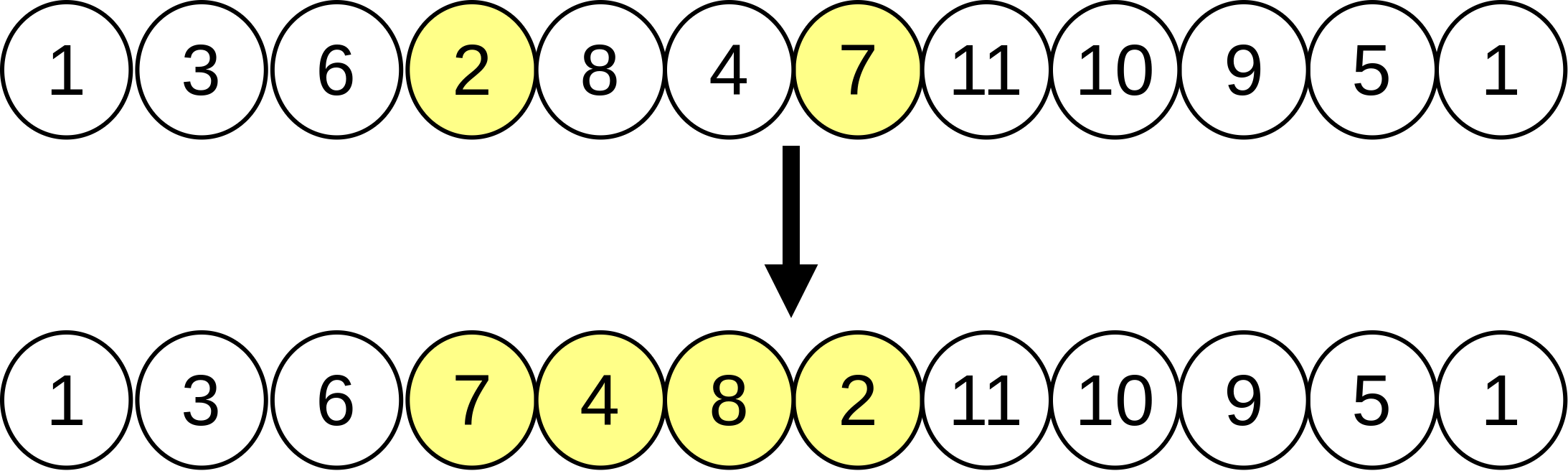}}
\caption{
Example of $2$-opt move with indices $(i=3,\ j=6)$, assuming zero-based numbering, where the position of all the nodes between the two indices is reversed.
}
\label{fig:2opt}
\end{figure}

\paragraph{Action.} Actions are elements within the neighbourhood defined by the operator $g$. For TSP, we consider $2$-opt moves~\cite{lin1973effective} that consist of selecting a tuple of indices $(i,\ j)$ and reversing the order of the nodes between $\delta_i$ and $\delta_j$. 
In Figure~\ref{fig:2opt} we illustrate an example of $2$-opt move with $(i=3,\ j=6)$, assuming zero-based numbering, where the order of all the nodes between $\delta_3=2$ and $\delta_6=7$ is reversed to obtain the new tour. 

\paragraph{Reward.} At each step $T$, the reward is computed as $r^T = L(\delta^{*T}) - L(\delta^{*T-1}$), hence the agent is rewarded only when improving on the best solution found.

The above elements with the state transitions derived by the operator neighbourhood define a Markov Decision Process (MDP)~\cite{bellman1957markovian,puterman1990markov} that we call improvement MDP that terminates after $T_{\text{max}}$ the number of steps in an episode.
Even though this work is focused on the Euclidean TSP and its variants, the described framework extends easily to other routing and CO problems.

\subsection{Search in Neural CO}
Search algorithms such as Beam Search (BS) and Monte Carlo Tree Search (MCTS)~\cite{coulom2006efficient} have been widely used in the literature on neural CO~\cite{joshi2019efficient,oren2021solo, choo2022simulation}.
Typically, they are used online with autoregressive constructive methods to boost their performance at inference time.
However, many of the search techniques that work well for constructive heuristics are difficult to extend efficiently to the improvement setting.
This is due to the fact that constructive methods work on a short horizon, i.e. the number of steps required to obtain a solution, which is defined by the number of variables in the problem.
On the contrary, improvement policies often require many more iterations to achieve good performance, see e.g.~\citet{ma2021learning}. 
Although MCTS has been widely applied in conjunction with DRL policies, yielding impressive results~\cite{silver2016mastering}, a notable drawback lies in the computational cost associated with its backpropagation procedure~\cite{choo2022simulation}. 
This limitation renders MCTS less suitable for the context of CO, particularly when dealing with large search spaces that are difficult to explore.
In the literature on neural CO, BS has emerged as a practical alternative to MCTS. This approach strikes a favourable balance between search capability and runtime complexity, making it a promising choice for addressing the challenges inherent in CO scenarios~\cite{vinyals2015pointer,nazari2018reinforcement,kool2018attention,joshi2019efficient}.

\paragraph{Adaptive Methods for Neural CO}
In recent developments within the field of neural CO, a novel trend has emerged in search methods that incorporate techniques for online adaptation of policy parameters during inference.
This trend finds inspiration in the work of~\citet{hottung2021efficient}, who introduced Efficient Active Search (EAS) as an enhanced version of Active Search (AS)\cite{bello2017neural}.
EAS focuses on training only a small subset of policy weights, significantly reducing its computational footprint.
Simulation Guided Beam Search (SGBS)~\cite{choo2022simulation} employs ``simulations'' (i.e. policy rollouts) to assess expanded nodes in BS and seamlessly integrates with EAS for online policy adaptation.
SGBS's lookahead capability facilitates informed node selection without the complexities associated with intricate backpropagation techniques as in MCTS.
However, it's important to note that SGBS samples from the DRL constructive policy until a leaf node is reached for node evaluation, rendering it too computationally intensive for improvement methods.
More recently, \citet{son2023meta} proposed Meta-SAGE that uses meta-learning and search in order to scale pre-trained models to large TSP instances. 
Introducing a bilevel formulation, the algorithm is made of two components: a scheduled adaptation with guided exploration (SAGE) that updates parameters at test time and a scale meta-learner that generates scale-aware context embeddings.

\section{Searching with LRBS}

\begin{figure}[ht]
\vskip 0.2in
\begin{center}
\centerline{\includegraphics[width=\columnwidth, scale=0.5]{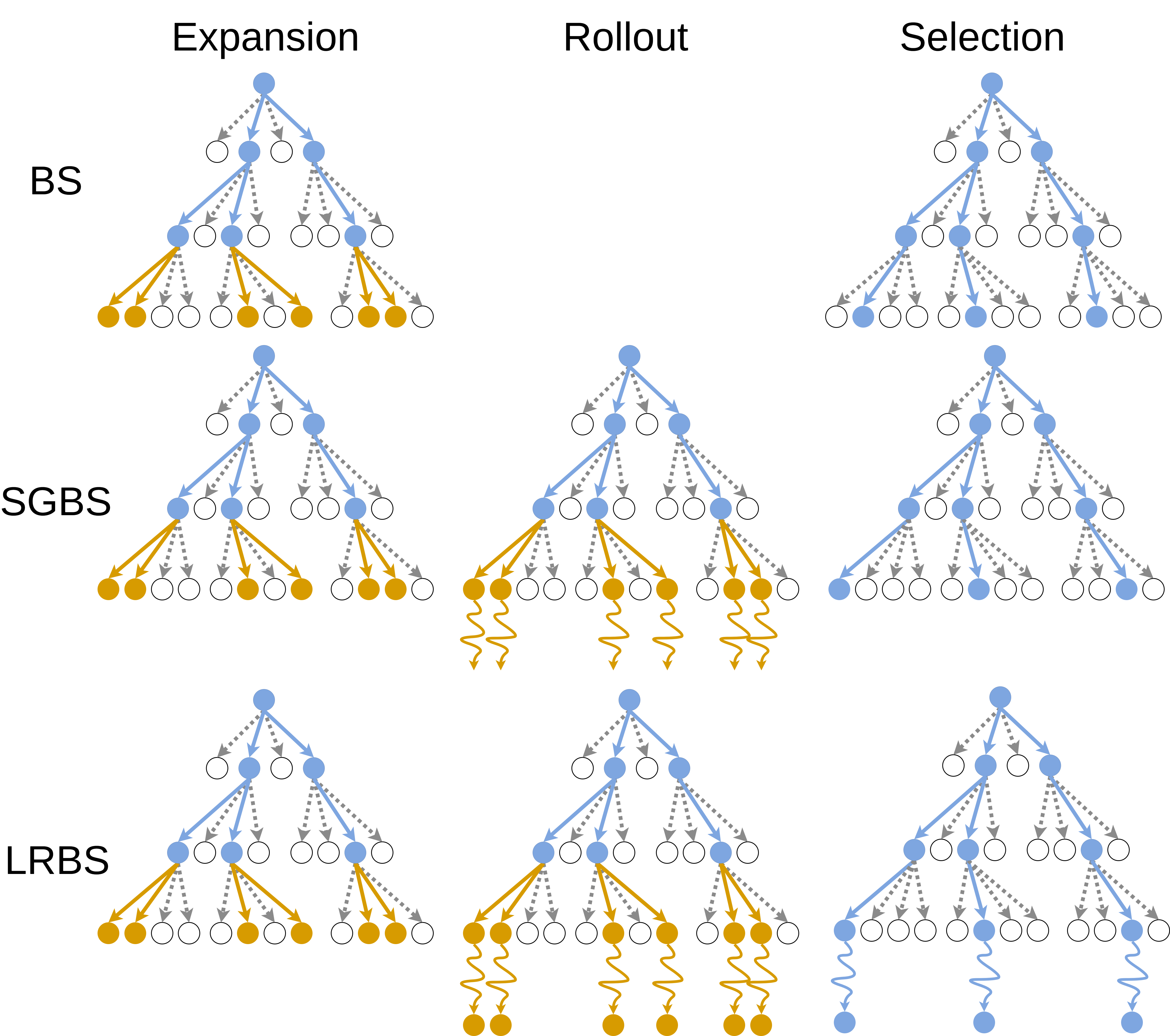}}
\caption{
Comparison of BS, SGBS, and LRBS across three algorithmic phases.
On the left, we illustrate the ``Expansion'' step, which shares similarities among the three algorithms.
Highlighted in blue are the $\beta$ paths of the active nodes within the beam and among their children the $\alpha$ yellow nodes represent the best ones selected for expansion. 
Starting from the selected children, SGBS and LRBS apply the DRL policy in the ``Rollout''. 
Finally at the ``Selection'' step the beam is updated and grown down the search tree. Illustration inspired by~\citet{choo2022simulation}.}
\label{fig:search}
\end{center}
\vskip -0.2in
\end{figure}

In this section, we describe our beam search strategy for CO improvement heuristics.
To overcome the limitations of previous methods in the improvement MDP, we propose an effective beam search approach that allows to trade-off between the additional computational cost of search and heuristic performance.
Additionally, our approach mitigates the effect of the longer episodic horizon in the improvement MDP by reducing the effective horizon on which the DRL policy works.

\subsection{The LRBS algorithm}

\begin{algorithm}[tb]
   \caption{Limited Rollout Beam Search}
   \label{alg:lrbs}
\begin{algorithmic}
   \STATE {\bfseries Input:} initial solution $\delta^0$, pre-trained policy $\pi$, parameters $(\alpha, \beta, n_s, T_{\text{max}})$, objective function $f$
   \STATE
   \STATE $\delta_{\text{best}}$ $\leftarrow \delta^0$
   \STATE $R \leftarrow \{\}$
   \STATE $B\leftarrow $ {\bfseries sample} $\alpha\times\beta$ tours from $\pi(\cdot|\delta^0)$ 
   \FOR{$\delta^1_i$ {\bfseries in} $B$}
       \STATE $\delta^{n_s}_i \leftarrow$ rollout $\pi$ for $n_s$ steps starting at $\delta^1_i$
       \STATE {\bfseries add} $\delta^{n_s}_i$ {\bfseries to} $R$
       \STATE {\bfseries update} $\delta_{\text{best}}$
   \ENDFOR
   \STATE $B \leftarrow$ select the best $\beta$ elements in $R$ according to $f$
   \STATE
   \STATE $t \leftarrow n_s$
   \WHILE{$t < T_{\text{max}}$}
   \STATE $R \leftarrow \{\}$
   \STATE $B\leftarrow $ {\bfseries for each} $\delta_i^t$  {\bfseries in} $B${ \bfseries sample} $\alpha$ tours from $\pi(\cdot|\delta_i^t)$
   \FOR{$\delta^t_i$ {\bfseries in} $B$}
       \STATE $\delta^{n_s + t}_i \leftarrow$ rollout $\pi$ for $n_s$ steps starting at $\delta^t_i$
       \STATE {\bfseries add} $\delta^{n_s + t}_i$ {\bfseries to} $R$
       \STATE {\bfseries update} $\delta_{\text{best}}$
   \ENDFOR
   \STATE $B \leftarrow$ select the best $\beta$ elements in $R$ according to $f$
   \STATE $t \leftarrow t + n_s$
   \ENDWHILE
   \STATE {\bfseries return} $\delta_{\text{best}}$
\end{algorithmic}
\end{algorithm}

Solving a CO problem with the DRL framework in Section~\ref{subsec:improv} can be seen as traversing a search tree using policy $\pi$ to decide the path to follow.
Nodes in the tree represent solutions to the problem, with the initial solution $\delta^0$ being the root node, and edges possible improvement actions (e.g. $2$-opt moves) that transform one solution into the other.
In Algorithm~\ref{alg:lrbs} we present LRBS, the algorithm starts at the root node and carries out its search down the tree in a breath-first fashion by keeping a beam of $\beta$ active nodes for each depth level and exploring $\alpha$ of their children, thus limiting the branching factor (see Figure~\ref{fig:search}).
Contrary to other search problems, there are no terminal nodes to reach in the improvement MDP of Section~\ref{subsec:improv}.
Hence, exploration is carried out until the explored paths in the search tree reach a fixed depth ($T_\text{max}$) and the best solution found is returned. 
In the LRBS algorithm, there are two main operations that can be described as follows.

\paragraph{Expansion and Rollout.} LRBS introduces into the standard BS expansion step a limited policy rollout.
Specifically, for each active node $o_k$ in the beam, $\alpha$ distinct children are sampled according to the probability distribution of $\pi(\cdot| o_k)$ (depicted in the left column in Figure~\ref{fig:search}) and then, from the resulting $\beta \times \alpha$ states, the policy is rolled out for $n_s$ steps to obtain solutions $o_{k+n_s}$ (as shown in the middle column of Figure~\ref{fig:search}).

\paragraph{Selection.} To update its beam, LRBS selects the best $\beta$ solutions according to the objective function $f$, e.g. $L$ in the improvement MDP described in Section~\ref{subsec:improv}, and then the search continues from the new resulting beam front (right column of Figure~\ref{fig:search}).  

In LRBS, the limited length rollouts have a considerable impact on the search capabilities of the algorithm.
Within the improvement MDP, the ability of the DRL agent to explore good solutions is highly constrained to the neighbourhood spanned by the used operator $g$ and the derived available actions.  
This implies that more than one step may be needed to reach a better solution than the current one, and even worse solutions may be observed in the path to an improved solution.
By incorporating $n_s$ steps of policy rollout before selection, instead of a single-step look-ahead as in BS, LRBS harnesses the improvement potential of $\pi$ and enhances its planning capabilities through exploratory actions facilitated by the beam.

\section{Adapting Pre-Trained Policies with LRBS}
While the search capabilities of LRBS mitigate the effect of distributional shifts when scaling to larger problem instances than those seen while training, its performance is limited by the pre-trained policy. 
In this section, we introduce an adaptive framework in which we combine LRBS with EAS to update the pre-trained DRL policy.
We study the effectiveness of this approach in two different scenarios: offline fine-tuning (FT) and online adaptation (OA).

In EAS, a small set of new parameters $\phi$ is introduced by adding a few layers into the agent's neural network, that in encoder-decoder architectures are usually placed in the final layers of the decoder.
To reduce the computational burden of previous adaptive methods, \citet{hottung2021efficient} proposed to only train the new weights $\phi$, making EAS extremely fast.
To update $\phi$ in constructive heuristics, EAS utilizes a loss function consisting of an RL component, aiming to reduce the cost of generated solutions, and an imitation learning component, which increases the probability of generating the best solution seen so far.
However, it is not straightforward to apply EAS in the improvement MDP since running multiple times the improvement heuristic for the total number of steps required to achieve a good solution and then adapting $\phi$ would incur extremely long computational times.
Instead, in LRBS we can incorporate easily EAS by updating the new weights on the limited rollouts used in node expansion.

To fine-tune the pre-trained policy, we assume a limited set ($\mathcal{S}_{FT}$) of instances in the target problem distribution is available and train $\phi$ to maximize the reward achieved over the LRBS rollouts with the RL loss function of EAS, leading to the gradient:
\begin{equation}
\label{eq:ftgrad}
    \nabla_\phi\mathcal{L}_R(\phi) = \mathbb{E}_\pi[(R(\delta_{\text{LRBS}}) - b)\nabla_\phi\log\pi_\phi(\delta_{\text{LRBS}})]
\end{equation}
where $\delta_{\text{LRBS}}$ is a rollout of $n_s$ steps and $b$ is a baseline (as in other works, we use the one proposed in~\citet{kwon2020pomo}).
This scenario is representative of many domains where similar CO problems have to be solved several times and past instances can be used for fine-tuning.
In our experiments, the instances in $\mathcal{S}_{FT}$ are solved only once by the LRBS algorithm and after each policy rollout the new parameters $\phi$ are updated according to the gradient in Equation~\ref{eq:ftgrad}.
Similarly, in the online adaptation scenario, we update the EAS weights at inference time with the approach described above.
However, the EAS parameters are reset before solving each batch of test problems, hence, the extra policy weights adapt solely to the instance being solved.

\section{Experimental Results}
\label{sec:exp_results}

In this section, we report experimental results on the search capabilities of LRBS and its effect on the generalization of pre-trained DRL agents to large TSP instances and two pickup and delivery variants.
In our study, we use checkpoints of models from~\citet{d2020learning}, pre-trained on Euclidean TSP instances with $100$ nodes, and from~\cite{ma2022efficient}, pre-trained on PDTSP and PDTSPL instances with $100$ nodes.
\citet{ma2021learning} recently proposed the Dual-Aspect Collaborative Transformer (DACT) architecture for the improvement of TSP solutions with $2$-opt moves.
Even though DACT performs better than the model from~\citet{d2020learning} in the authors' study, the latter architecture showed much better scalability in our preliminary investigations and even outperformed DACT when both were coupled with LRBS.

In all our experiments on TSP, for LRBS, we set a ``search budget'' such that $\alpha\times\beta = 60$ and fix the other parameters to $n_s = 20$ and $T_\text{max}=5000$.
On PDTSP and PDTSPL we reduce the budget to $40$ and when doing adaptation we use $n_s=10$ to lower memory consumption. 
The best values of $\alpha$ and $\beta$ for each dataset were determined by testing the method on a set of $10$ randomly generated instances of the same size as those in the test set.

\paragraph{Tests datasets.} 
The TSP instances in our experiments are generated according to the method in~\citet{kool2018attention} where the coordinates of nodes are sampled uniformly at random in the unit square.
We consider problems with $N = \{100, 150, 200, 500, 1000\}$ nodes.
To ensure a fair comparison with the pre-trained policies, for $N=100$ we use the same $10,000$ test instances of~\citet{d2020learning}.
For the other problems, we generate datasets with $1,000$ random instances for $N=\{125, 200\}$ and with $128$ instances for $N={500, 1000}$. 
For PDTSP and PDTSPL experiments we generate sets of $128$ random instances with $200$ and $500$ nodes.  
In the following, we refer to the test dataset with problems with $N$ nodes as TSP$N$, PDTSP$N$ and PDTSPL$N$, respectively.

\paragraph{Baselines.}
To assess the performance of our method, we compare it with the pre-trained policy of~\citet{d2020learning} and DACT, also with $8$x of the augmentations introduced in~\citet{kwon2020pomo} (A-DACT).
Moreover, among the baselines, we include a modification on SGBS (SGBS+C) with limited rollouts (as in LRBS) to work with improvement heuristics, e.g. the pre-trained DRL policy of~\citet{d2020learning}.
The search approach is similar to LRBS but the new beam front is selected from the direct children of the previous beam front, based on the information of the limited rollouts. 
Finally, we report the performance of recent constructive approaches and related algorithms that use search and adaptation to gauge the gap with improvement techniques~\cite{kwon2020pomo, hottung2021efficient, choo2022simulation, son2023meta}.
Recent methods~\cite{drakulic2024bq,sun2023difusco,luo2024neural} achieve better generalization than the considered baseline on TSP.
However, such methods typically require specialized policy training and cannot be directly used on any pre-trained policy as we do in this work.

\paragraph{Computational resources.} 
We run all our experiments using a single NVIDIA Ampere GPU with 64GB of HBM2 memory. 

\begin{table}[ht!]
\setlength{\tabcolsep}{1mm}
\fontsize{9pt}{9pt}\selectfont
\centering
\begin{sc}
\begin{tabular}{l|ccc|ccr}
\toprule
\multicolumn{1}{c}{}&\multicolumn{3}{c}{$N=100$}&\multicolumn{3}{c}{$N=150$}\\
Method&Obj.&Gap&Time&Obj.&Gap&Time\\
\midrule
Concorde&$7.75$&$0.0\%$&-&$9.35$&$0.0\%$&-\\
\midrule
POMO\footnotemark[1]&$7.77$&$0.078\%$&$3$h&$9.37$&$0.33\%$&$1$h\\ 
SGBS\footnotemark[1]&$7.76$&$0.058\%$&$0.2$h&$9.36$&$0.22\%$&$0.1$h\\ 
EAS\footnotemark[1]&$7.76$&$0.044\%$&$15$h&$9.35$&$0.12\%$&$10$h\\ 
SGBS+EAS\footnotemark[1]&$7.76$&$0.024\%$&$15$h&$\textbf{9.35}$&$\textbf{0.08\%}$&$10$h\\ 
\midrule
DACT [$10$k]&$7.79$&$0.463\%$&$1.7$h&$10.20$&$9.12\%$&$0.4$h\\
A-DACT [$10$k]&$7.76$&$0.101\%$&$13$h&$9.86$&$5.54\%$&$2.9$h\\
Costa&$7.76$&$0.065\%$&$19$h&$9.37$&$0.25\%$&$3.2$h\\
SGBS+C [$5$k]&$7.78$&$0.335\%$&$193$h&$9.44$&$1.00\%$&$31$h\\
Beam S. [$5$k]&$7.76$&$0.015\%$&$19$h&$9.36$&$0.18\%$&$6.4$h\\
\midrule
Ours [$5$k]&$7.76$&$0.014\%$&$19$h&$9.36$&$0.16\%$&$3.2$h\\
Ours+OA [$5$k]&$\textbf{7.76}$&$\textbf{0.013\%}$&$22$h&$9.36$&$0.13\%$&$3.6$h\\
\bottomrule
\end{tabular}
\end{sc}
 \parbox{\linewidth}{\small 
    \footnotemark[1] The results (i.e., obj. values, gaps, and time) are adopted from~\cite{choo2022simulation}.
  }
  \caption{Performance evaluation on TSP$100$ and TSP$150$ test datasets. For each considered method and dataset, we report the average objective value (Obj.), the average optimality gap (Gap) and the total time required to solve the instances in the test set (Time). The best-performing method for each dataset is highlighted in bold.}
\label{tab:exp1}
\end{table}

\subsection{Boosting In-Distribution Performance}
In Table~\ref{tab:exp1} we report the results of LRBS on test instances with $100$ and $150$ that are close to the training data distribution.
Our method outperforms all the considered baselines on TSP$100$ and has the best results among improvement heuristics on TSP$150$.
On TSP$150$ the constructive baselines show slightly better gaps than LRBS, but our approach has considerably lower runtime.
From our analysis, on test instances close to the training data distribution the best LRBS configuration is $(\beta=60,\ \alpha=1)$.
While such a configuration corresponds to $\beta$ parallel runs of the policy, introducing limited rollouts allows us to perform online adaption and achieve improved performance. 

\begin{table*}
\begin{center}
\begin{small}
\begin{sc}
\begin{tabular}{l|ccc|ccc|ccr}
\toprule
\multicolumn{1}{c}{}&\multicolumn{3}{c}{TSP$200$}&\multicolumn{3}{c}{TSP$500$}&\multicolumn{3}{c}{TSP$1000$}\\
Method&Obj.&Gap&Time&Obj.&Gap&Time&Obj.&Gap&Time\\
\midrule
Concorde&$10.704$&$0.0\%$&-&$16.530$&$0.0\%$&-&$23.144$&$0.0\%$&-\\
\midrule
EAS\footnotemark[2]&$10.736$&$0.455\%$&$2.4$h&$18.135$&$9.362\%$&$4.3$h&$30.744$&$32.869\%$&$20$h\\
SGBS+EAS\footnotemark[2]&$10.734$&$0.436\%$&$2.1$h&$18.191$&$9.963\%$&$4.2$h&$28.413$&$22.795\%$&$19$h\\
Meta-SAGE\footnotemark[2]&$\textbf{10.729}$&$\textbf{0.391\%}$&$2.1$h&$17.131$&$3.559\%$&$3.8$h&$25.924$&$12.038\%$&$18$h\\

\midrule
DACT [$10$k]&$15.450$&$34.346\%$&$0.6$h& $154.339$&$833\%$&$0.4$h&$421.76$&$1722\%$&$1.8$h\\
A-DACT [$10$k]&$14,345$&$34.023\%$&$4.4$h& $147.127$&$790\%$&$3.3$h&$412,787$&$1683\%$&$11.9$h\\
Costa&$10.817$&$1.063\%$&$1.6$h& 
$18.078$&$9.361\%$&$0.8$h&
$30.655$&$32.460\%$&$3.1$h\\
Costa&$10.789$&$0.796\%$&$4.1$h&
$17.971$&$8.717\%$&$1.9$h&
$30.439$&$31.526\%$&$7.8$h\\
SGBS+C [$2$k]&$10.929$&$2.100\%$&$8.5$h&$20.690$&$25.169\%$&$7.6$h&$60.518$&$161.5\%$&$31.4$h\\
SGBS+C [$5$k]&$10.903$&$1.858\%$&$21.2$h&$18.455$&$11.642\%$&$19.1$h&$47.083$&$103.44$&$78.6$h\\
Beam S. [$5$k]&$11.137$&$4.051\%$&$6.4$h&$17.851$&$7.993\%$&$2.5$h&
$26.507$&$14.536\%$&$8.7$h\\
\midrule
Ours [$2$k]&$10.782$&$0.738\%$&$1.6$h&
$17.684$&$6.902\%$&$0.8$h&
$32.368$&$39.970\%$&$3.1$h\\
Ours [$5$k]&$10.771$&$0.633\%$&$4.1$h& $17.309$&$4.633\%$&$1.9$h& $27.922$&$20.740\%$&$7.8$h\\
Ours+OA [$2$k]&$10.771$&$0.629\%$&$1.9$h&
$17.443$&$5.523\%$&$0.9$h&
$28.468$&$23.008\%$&$3.3$h\\
Ours+OA [$5$k]&$10.760$&$0.528\%$&$4.8$h&
$\textbf{17.187}$&$\textbf{3.973\%}$&$2.0$h&
$\textbf{25.895}$&$\textbf{11.889\%}$&$8.1$h\\
\midrule
\midrule
Ours+FT [$2$k]&$10.768$&$0.599\%$&$2.6$h&
$17.303$&$4.680\%$&$0.9$h&
$28.891$&$24.838\%$&$3.9$h\\
Ours+FT [$5$k]&$10.757$&$0.504\%$&$4.6$h&
$17.102$&$3.463\%$&$2.0$h&
$25.801$&$11.483\%$&$8.6$h\\
\bottomrule
\end{tabular}
\end{sc}
\end{small}
\end{center}
 \parbox{\linewidth}{\small 
    \footnotemark[2] The results (i.e., obj. values, gaps, and time) are adopted from~\cite{son2023meta} where the TSP$500$ and TSP$1000$ datasets have $128$ instances.
  }
\caption{Performance evaluation on TSP$200$, TSP$500$ and TSP$1000$ test datasets. For each considered method and dataset, we report the average objective value (Obj.), the average optimality gap (Gap) and the total time required to solve the instances in the test set (Time). The best-performing method for each dataset is highlighted in bold.}
\label{tab:exp2}
\end{table*}

\subsection{Out-of-Distribution Exploration with a Pre-Trained Policy}
In the first part of Table~\ref{tab:exp2} we show results on the generalization power of LRBS when solving larger TSP problems with $200$, $500$ and $1000$  nodes.
The configurations used for LRBS are $(\beta = 30,\ \alpha = 2)$, $(\beta=15,\ \alpha=4)$ and $(\beta=5,\ \alpha=12)$ for the TSP$200$, TSP$500$ and TSP$1000$, respectively.
As the test set distribution shifts away from the training distribution we observe that increasing the number of children evaluated for each node in the beam front improves on generalization.
While on smaller instances the policy can select good actions and more exploitation with lower $\alpha$ leads to the best performance, on larger instances increasing $\alpha$ allows to compensate for the imprecision of the agent and yields better results.   
On these test datasets, LRBS scales better than other improvement heuristics achieving optimality gaps close to those of constructive approaches.
From our experiments, we can observe that the augmentations employed by~\citet{ma2021learning} improve considerably the policy performance on instances with the same size as the training set.
However, when considering larger graphs the benefit of the augmentations becomes less pronounced and the algorithm fails to scale.
On the contrary, online exploration with LRBS mitigates the performance degeneration due to distributional shift and our method even improves on the results that the policy of~\citet{d2020learning} would achieve if exploring the solution space for the same time as LRBS and using on average $12$x more $2$-opt operations. 
On larger instances, LRBS is not competitive with Meta-SAGE but achieves optimality gaps comparable to those of EAS and SGBS+EAS even improving on their performance as the instances get larger.

Turning our attention to the comparison of LRBS to BS, the results in Table~\ref{tab:exp2} present an interesting phenomenon.
On the smaller instances with up to $500$ nodes, LRBS is faster and achieves much lower optimality gaps, even $6$x smaller than BS.
However, on the largest problems of the TSP$1000$ dataset BS has superior performance than LRBS.
This result strongly suggests that as the distributional shift between the training and test instances gets very large the step-wise selection process of BS is better than the rollouts of LRBS in limiting the performance degradation of the policy.
This further motivates the need for adaptive strategies to overcome the limitations posed by the pre-trained model.

\subsection{Generalization via Adaptation}
In the second part of Table~\ref{tab:exp2}, we show results on the generalization of LRBS after fine-tuning the DRL policy on a small set of randomly generated instances with the same number of nodes as the test set (FT) and when adapting the policy parameters online (OA).
The LRBS configurations are the same used for the non-adaptive experiments, with the only exception of the LRBS + FT on the TSP$1000$ where we use $(\beta=10,\ \alpha=6)$.
For all the considered problems, the FT dataset of randomly generated instances is of size equal to $10\%$ of the test set size and each instance is solved only once using LRBS, running times include also the fine-tuning phase.
While LRBS+FT shows the best results for TSP$500$ and TSP$1000$, we do not highlight them in bold to keep a fair comparison with the baselines.
Our results show that fine-tuning on a limited set of problems allows LRBS to improve considerably on larger instances surpassing all the baselines on the TSP$500$ and TSP$1000$ benchmarks, while on the TSP$200$ the performance of LRBS is close to that of EAS. 
Even though online adaptation is less effective than fine-tuning, since policy weights are trained only on the instance being solved and reset thereafter, it achieves competitive results on the TSP$200$ and TSP$500$ datasets while outperforming constructive baselines on the TSP$1000$ instances.
Although FT achieves lower gaps, we highlight these results in Table~\ref{tab:exp2} for a fair comparison with the baselines.
These results show that introducing an adaptive component in the search process of LRBS can overcome the limitations posed by the adopted pre-trained policy.
In particular, on the larger TSP$1000$ problems, the use of LRBS alone fails to achieve the performance of the constructive baselines while both the offline and online adaptive approaches we propose almost halve the optimality gap of LRBS alone and even improve on the baselines.

\paragraph{Ablation study.}

\begin{table}[ht!]
\begin{center}
\begin{small}
\begin{sc}
\begin{tabular}{lccr}
\toprule
&TSP$200$&TSP$500$&TSP$1000$\\
\midrule
Ours+FT & $0.50\%$& $3.46\%$& $11.48\%$\\
w/o LRBS FT & $0.52\%$& $8.43\%$& $81.63\%$\\
w/o exp. & $1.44\%$& $7.57\%$& $16.06\%$\\
\bottomrule
\end{tabular}
\end{sc}
\end{small}
\end{center}
\caption{Ablation study on TSP$200$, TSP$500$ and TSP$1000$ datasets. For each algorithmic configuration, we report optimality gaps.}
\label{tab:ablation}
\end{table}

To evaluate the importance of each component in the fine-tuning experiments we conduct on LRBS, in Table~\ref{tab:ablation} we report the results of an ablation study by removing LRBS from the fine-tuning and inference phases.
In the first case (w/o LRBS FT in Table~\ref{tab:ablation}), the training framework of~\citet{d2020learning} is used to fine-tune the policy while in the latter (w/o Exp. in Table~\ref{tab:ablation}) we sample from the policy for the same time as the LRBS runtime.
To provide a fair comparison, when fine-tuning is done without LRBS the training runs for $n_s\times \beta\time \alpha$ steps to train on the same number of environment interactions.

\begin{table*}
\setlength{\tabcolsep}{1mm}
\fontsize{9pt}{9pt}\selectfont
\vskip 0.15in
\begin{center}
\begin{sc}
\begin{tabular}{l|ccc|ccc|ccc|ccr}
\toprule
\multicolumn{1}{c}{}&\multicolumn{3}{c}{$\text{PDTSP}200$}&\multicolumn{3}{c}{$\text{PDTSP}500$}&\multicolumn{3}{c}{$\text{PDTSPL}200$}&\multicolumn{3}{c}{$\text{PDTSPL}500$}\\
Method&Obj.&Gap&Time&Obj.&Gap&Time&Obj.&Gap&Time&Obj.&Gap&Time\\
\midrule
LKH&$12.913$&$0.0\%$&$3.4$h&$20.332$&$0.0\%$&$20.8$h&$29.322$&$0.0\%$&$2.7$h&$69.922$&$0.0\%$&$23.1$h\\
\midrule
N2S-A [$1$k]&$15.321$&$18.655\%$&$2.7$h&$114.789$&$464.668\%$&$34.6$h&$31.060$&$5.937\%$&$3.4$h&$185.862$&$165.900\%$&$51.0$h\\
N2S-A [$2$k]&$14.875$&$15.198\%$&$5.5$h&$95.513$&$369.723\%$&$69.3$h&$30.278$&$3.272\%$&$6.8$h&$176.543$&$152.583\%$&$102.3$h\\
N2S-A [$3$k]&$14.716$&$13.960\%$&$8.3$h&$89.731$&$341.330\%$&$103.9$h&$30.028$&$2.417\%$&$10.1$h&$173.354$&$148.017\%$&$153.5$h\\
\midrule
Ours [$1$k]&$14.710$&$13.918\%$&$1.0$h&$48.907$&$140.604\%$&$5.8$h&$30.228$&$3.103\%$&$1.4$h&$142.341$&$103.669\%$&$8.8$h\\
Ours [$2$k]&$14.433$&$11.773\%$&$2.0$h&$35.891$&$76.558\%$&$11.6$h&$29.869$&$1.874\%$&$2.8$h&$108.413$&$55.087\%$&$17.6$h\\
Ours [$3$k]&$14.320$&$10.899\%$&$3.0$h&$33.723$&$65.889\%$&$17.3$h&$29.721$&$1.370\%$&$4.2$h&$85.762$&$22.663\%$&$26.5$h\\
\midrule
\midrule
Ours+OA [$1$k]&$14.456$&$11.951\%$&$1.9$h&$39.885$&$96.201\%$&$11.8$h&$30.164$&$2.884\%$&$2.8$h&$108.849$&$55.726\%$&$21.6$h\\
Ours+OA [$2$k]&$14.161$&$9.664\%$&$3.7$h&$33.587$&$65.227\%$&$23.6$h&$29.839$&$1.774\%$&$5.6$h&$81.044$&$15.942\%$&$43.0$h\\
Ours+OA [$3$k]&$\textbf{13.987}$&$\textbf{8.321\%}$&$5.5$h&$\textbf{31.820}$&$\textbf{56.532\%}$&$35.4$h&$\textbf{29.716}$&$\textbf{1.358\%}$&$8.5$h&$\textbf{77.012}$&$\textbf{10.161\%}$&$64.2$h\\

\bottomrule
\end{tabular}
\end{sc}
\end{center}
\caption{Performance evaluation on PDTSP$200$, PDTSP$500$, PDTSPL$200$ and PDTSPL$500$ test datasets. For each considered method and dataset, we report the average objective value (Obj.), the average optimality gap (Gap) and the total time required to solve the instances in the test set (Time). The best-performing method for each dataset is highlighted in bold.}
\label{tab:exp_pdp}
\end{table*}

While on the TSP$200$ dataset removing LRBS from fine-tuning yields a gap close to that with LRBS, on TSP$500$ and TSP$1000$ there is a considerable performance degeneration.
On the contrary, when online exploration is replaced by sampling there is a much smaller effect on the generalization abilities of the model in the larger datasets but a greater decrease in performance in the TSP$200$ dataset.
This shows the strength of LRBS in the fine-tuning phase where exploration allows the policy to better adapt to larger instances, especially for larger instances where the policy can easily get stuck in local optima.
Moreover, for smaller problems, we observe that exploration in the fine-tuning phase is less critical but it has a considerable impact when applied online.

\paragraph{Computational efficiency}
From Table~\ref{tab:exp2} we can also observe that LRBS and its online adaptive variant not only present a lower runtime than the constructive methods but also scale better, i.e. the relative increase in runtime as the test problems get larger is smaller for our algorithms. 
The reason for this fact is the autoregressive nature of constructive heuristics. 
With improvement approaches, we keep a fixed number of steps hence the computational cost grows only due to the larger problems to be processed by the neural policy. 
However, constructive methods, by design, need to perform increasingly more steps to generate solutions for larger instances, thus incurring an additional computational burden as the size of the problems increases

\subsection{Pickup and Delivery Problems}
The pickup and delivery variant of TSP (PDTSP) consists of $n$ one-to-one pickup-delivery requests, meaning that there are $n$ pickup nodes with goods that need to be transported to $n$ corresponding delivery nodes. 
The objective is to find the shortest Hamiltonian cycle under the \emph{precedence} constraint that every pickup node has to be visited before its corresponding delivery node.
We also study PDTSP with the \emph{last-in-first-out} constraint (PDTSPL) that enforces a stack ordering between collected goods and delivery is allowed only for the good at the top of the stack.
For these problems,~\citet{ma2022efficient} define a \emph{removal-reinsertion} operator that selects a pickup-delivery request nodes $(\delta_{i^+},\ \delta_{i^-})$, positions $(j,\ k)$ and places node $\delta_{i^+}$ after node $\delta_j$ and node $\delta_{i^-}$ after $\delta_k$.

In Table~\ref{tab:exp_pdp} we report the results of applying LRBS on model checkpoints from~\citet{ma2022efficient} (N2S-A), pre-trained on pickup and delivery instances of size $100$, when solving PDTSP and PDTSPL instances with $N=200$ and $500$ nodes.
In these experiments, for N2S-A we use the same exploration strategy adopted by the authors where at inference time each instance solved is transformed into $\frac{1}{2}|N|$ different ones, using the augmentations of~\cite{kwon2020pomo}, and the policy is rolled out from each new instance.
Our results show that the online exploration approach of LRBS is much more effective than N2S-A when generalizing to larger instances. 
Not only in terms of pure performance but also computational efficiency.
On the smaller instances with $200$ nodes, LRBS achieves a good reduction of optimality gaps requiring less time than N2S-A even when performing online adaptation.
The PDTSP$500$ benchmark results are not satisfactory with optimality gaps well above $50\%$ but still, LRBS shows improved generalization compared to N2S-A reducing its gap by almost $6x$.
On the much more constrained PDTSPL$500$ problems instead, online search through LRBS outperforms N2S-A with a gap reduction close to $10x$ when adaption is employed.
Overall, the results of Table~\ref{tab:exp_pdp} are still far from being competitive with traditional solvers such as LKH but show the generalization potential of pre-trained policy with online search and adaptation.

\section{Conclusion}
In this study, we have introduced LRBS, a novel beam search method designed to complement DRL-based improvement heuristics for combinatorial optimization problems enhancing inference time performance and generalization.
LRBS offers a tailored approach that enables pre-trained models to efficiently handle problem instances of significantly larger scales, up to ten times bigger than those encountered during the DRL policy initial training phase.
To further enhance the generalization of pre-trained models, we integrate LRBS with EAS in offline and online adaptive scenarios. 
Our experimental evaluation shows LRBS's superiority over existing DRL improvement methods in the context of solving the Euclidean TSP and two pickup and delivery variants.
LRBS consistently outperforms alternative approaches proposed both for constructive and improvement heuristics.
Moreover, in our analysis, LRBS exhibits superior runtime efficiency when scaling to larger instances compared to established constructive baselines, showing how improvement heuristics coupled with adaptive and search approaches can be a viable alternative to constructive methods.

\bibliography{aaai25}

\appendix
\section{Sensitivity Analysis: Performance of Different LRBS Configurations}
\label{sec:app_b}
In Section~\ref{sec:exp_results} we report the results of LRBS with the best-performing parameter configurations, here we show its performance (without online adaptation) for different values of the parameters $\alpha$ and $\beta$ on  TSP datasets. For TSP$500$ and TSP$1000$ the datasets considered have $1000$ and $200$ instances, respectively, instead of $128$ as in the main results. Thus, computational times increase accordingly.

\begin{table}[H]\setlength{\tabcolsep}{1mm}
\fontsize{9pt}{9pt}\selectfont
\caption{Performance of different configurations of LRBS on TSP$100$ dataset.}
\vskip 0.15in
\begin{center}
\begin{small}
\begin{sc}
\begin{tabular}{rrll|rrrrrrr}
\toprule
$T_{\text{max}}$ &  $\beta$ &  $n_s$ & $\alpha$ &   Obj. &    Gap &   Obj. &   Time (h) \\
\midrule
 500 &  20 &            20 &           3 & 7.771 &  0.146 & 7.760 &  1.870 \\
 500 &  30 &            20 &           2 & 7.767 &  0.094 & 7.760 &  1.885 \\
 500 &  60 &            20 &           1 & 7.769 &  0.113 & 7.760 &  1.937 \\
 \midrule
1000 &  20 &            20 &           3 & 7.769 &  0.119 & 7.760 &  3.740 \\
1000 &  30 &            20 &           2 & 7.765 &  0.073 & 7.760 &  3.771 \\
1000 &  60 &            20 &           1 & 7.764 &  0.049 & 7.760 &  3.874 \\
 \midrule
2000 &  20 &            20 &           3 & 7.768 &  0.107 & 7.760 &  7.481 \\
2000 &  30 &            20 &           2 & 7.765 &  0.063 & 7.760 &  7.542 \\
2000 &  60 &            20 &           1 & 7.762 &  0.027 & 7.760 &  7.748 \\
 \midrule
5000 &  20 &            20 &           3 & 7.767 &  0.093 & 7.760 & 18.701 \\
5000 &  30 &            20 &           2 & 7.764 &  0.056 & 7.760 & 18.853 \\
5000 &  60 &            20 &           1 & 7.761 &  0.015 & 7.760 & 19.370 \\
\bottomrule
\end{tabular}
\end{sc}
\end{small}
\end{center}
\end{table}

\begin{table}[H]\setlength{\tabcolsep}{1mm}
\fontsize{9pt}{9pt}\selectfont
\caption{Performance of different configurations of LRBS on TSP$150$ dataset.}
\vskip 0.15in
\begin{center}
\begin{small}
\begin{sc}
\begin{tabular}{lrrllrrrrrrr}
\toprule
$T_{\text{max}}$ &  $\beta$ &  $n_s$ &  $\alpha$ &   Obj. &    Gap &   Obj. &   Time (h) \\
  \midrule
  500 &          20 &            20 &           3 & 9.412 & 0.649 & 9.351 & 0.309 \\
  500 &          30 &            20 &           2 & 9.401 & 0.538 & 9.351 & 0.312 \\
  500 &          60 &            20 &           1 & 9.440 & 0.950 & 9.351 & 0.321 \\
  \midrule
 1000 &          20 &            20 &           3 & 9.396 & 0.487 & 9.351 & 0.619 \\
 1000 &          30 &            20 &           2 & 9.384 & 0.357 & 9.351 & 0.624 \\
 1000 &          60 &            20 &           1 & 9.395 & 0.475 & 9.351 & 0.641 \\
  \midrule
 2000 &          20 &            20 &           3 & 9.389 & 0.413 & 9.351 & 1.238 \\
 2000 &          30 &            20 &           2 & 9.379 & 0.300 & 9.351 & 1.247 \\
 2000 &          60 &            20 &           1 & 9.377 & 0.279 & 9.351 & 1.282 \\
  \midrule
 5000 &          20 &            20 &           3 & 9.384 & 0.358 & 9.351 & 3.094 \\
 5000 &          30 &            20 &           2 & 9.374 & 0.253 & 9.351 & 3.116 \\
 5000 &          60 &            20 &           1 & 9.366 & 0.163 & 9.351 & 3.206 \\
\bottomrule
\end{tabular}
\end{sc}
\end{small}
\end{center}
\end{table}

\begin{table}[H]\setlength{\tabcolsep}{1mm}
\fontsize{9pt}{9pt}\selectfont
\caption{Performance of different configurations of LRBS on TSP$200$ dataset.}
\vskip 0.15in
\begin{center}
\begin{small}
\begin{sc}
\begin{tabular}{lrrllrrrrrrr}
\toprule
$T_{\text{max}}$ &  $\beta$ &  $n_s$ &  $\alpha$ &   Obj. &    Gap &   Obj. &   Time (h) \\
\midrule
 500 &          20 &            20 &           3 & 10.888 & 1.720 & 10.703 & 0.408 \\
 500 &          30 &            20 &           2 & 10.887 & 1.714 & 10.703 & 0.412 \\
 500 &          60 &            20 &           1 & 11.018 & 2.942 & 10.703 & 0.422 \\
\midrule
1000 &          20 &            20 &           3 & 10.817 & 1.064 & 10.703 & 0.815 \\
1000 &          30 &            20 &           2 & 10.803 & 0.932 & 10.703 & 0.823 \\
1000 &          60 &            20 &           1 & 10.868 & 1.532 & 10.703 & 0.844 \\
\midrule
2000 &          20 &            20 &           3 & 10.796 & 0.860 & 10.703 & 1.631 \\
2000 &          30 &            20 &           2 & 10.783 & 0.738 & 10.703 & 1.646 \\
2000 &          60 &            20 &           1 & 10.811 & 1.004 & 10.703 & 1.689 \\
\midrule
5000 &          20 &            20 &           3 & 10.781 & 0.723 & 10.703 & 4.077 \\
5000 &          30 &            20 &           2 & 10.771 & 0.633 & 10.703 & 4.116 \\
5000 &          60 &            20 &           1 & 10.774 & 0.655 & 10.703 & 4.222 \\
\bottomrule
\end{tabular}
\end{sc}
\end{small}
\end{center}
\end{table}

\begin{table}[H]\setlength{\tabcolsep}{1mm}
\fontsize{9pt}{9pt}\selectfont
\caption{Performance of different configurations of LRBS on TSP$500$ dataset.}
\vskip 0.15in
\begin{center}
\begin{small}
\begin{sc}
\begin{tabular}{lrrllrrrrrrr}
\toprule
$T_{\text{max}}$ &  $\beta$ &  $n_s$ &  $\alpha$ &   Obj. &    Gap &   Obj. &   Time (h) \\
\midrule
500 &           6 &            20 &          10 & 27.302 & 65.174 & 16.530 &  1.465 \\
500 &          10 &            20 &           6 & 27.572 & 66.806 & 16.530 &  1.470 \\
500 &          15 &            20 &           4 & 27.934 & 68.999 & 16.530 &  1.479 \\
500 &          20 &            20 &           3 & 28.231 & 70.798 & 16.530 &  1.486 \\
500 &          30 &            20 &           2 & 28.812 & 74.308 & 16.530 &  1.501 \\
500 &          60 &            20 &           1 & 30.832 & 86.532 & 16.530 &  1.544 \\
\midrule
1000 &           6 &            20 &          10 & 18.772 & 13.564 & 16.530 &  2.930 \\
1000 &          10 &            20 &           6 & 18.866 & 14.130 & 16.530 &  2.940 \\
1000 &          15 &            20 &           4 & 19.002 & 14.954 & 16.530 &  2.958 \\
1000 &          20 &            20 &           3 & 19.134 & 15.753 & 16.530 &  2.971 \\
1000 &          30 &            20 &           2 & 19.441 & 17.612 & 16.530 &  3.002 \\
1000 &          60 &            20 &           1 & 20.885 & 26.346 & 16.530 &  3.087 \\
\midrule
2000 &           6 &            20 &          10 & 17.636 &  6.688 & 16.530 &  5.861 \\
2000 &          10 &            20 &           6 & 17.638 &  6.700 & 16.530 &  5.879 \\
2000 &          15 &            20 &           4 & 17.668 &  6.880 & 16.530 &  5.916 \\
2000 &          20 &            20 &           3 & 17.703 &  7.093 & 16.530 &  5.942 \\
2000 &          30 &            20 &           2 & 17.832 &  7.876 & 16.530 &  6.004 \\
2000 &          60 &            20 &           1 & 18.814 & 13.813 & 16.530 &  6.175 \\
\midrule
5000 &           6 &            20 &          10 & 17.293 &  4.613 & 16.530 & 14.661 \\
5000 &          10 &            20 &           6 & 17.292 &  4.608 & 16.530 & 14.698 \\
5000 &          15 &            20 &           4 & 17.296 &  4.630 & 16.530 & 14.790 \\
5000 &          20 &            20 &           3 & 17.306 &  4.691 & 16.530 & 14.854 \\
5000 &          30 &            20 &           2 & 17.361 &  5.028 & 16.530 & 15.010 \\
5000 &          60 &            20 &           1 & 18.250 & 10.404 & 16.530 & 15.437 \\
\bottomrule
\end{tabular}
\end{sc}
\end{small}
\end{center}
\end{table}

\begin{table}[H]\setlength{\tabcolsep}{1mm}
\fontsize{9pt}{9pt}\selectfont
\caption{Performance of different configurations of LRBS on TSP$1000$ dataset.}
\vskip 0.15in
\begin{center}
\begin{small}
\begin{sc}
\begin{tabular}{lrrllrrrrrrr}
\toprule
$T_{\text{max}}$ &  $\beta$ &  $n_s$ &  $\alpha$ &   Obj. &    Gap &   Obj. &   Time (h) \\
\midrule
 500 &           5 &            20 &          12 & 100.261 & 333.244 & 23.144 &  1.203 \\
 500 &           6 &            20 &          10 & 100.528 & 334.397 & 23.144 &  1.205 \\
 500 &          10 &            20 &           6 & 101.007 & 336.469 & 23.144 &  1.209 \\
 500 &          15 &            20 &           4 & 101.849 & 340.108 & 23.144 &  1.216 \\
 500 &          20 &            20 &           3 & 102.719 & 343.869 & 23.144 &  1.223 \\
 500 &          30 &            20 &           2 & 104.073 & 349.718 & 23.144 &  1.233 \\
 500 &          60 &            20 &           1 & 108.885 & 370.514 & 23.144 &  1.270 \\
 \midrule
1000 &           5 &            20 &          12 &  51.494 & 122.508 & 23.144 &  2.406 \\
1000 &           6 &            20 &          10 &  51.751 & 123.622 & 23.144 &  2.410 \\
1000 &          10 &            20 &           6 &  52.408 & 126.460 & 23.144 &  2.418 \\
1000 &          15 &            20 &           4 &  53.162 & 129.720 & 23.144 &  2.431 \\
1000 &          20 &            20 &           3 &  53.843 & 132.664 & 23.144 &  2.445 \\
1000 &          30 &            20 &           2 &  54.993 & 137.629 & 23.144 &  2.466 \\
1000 &          60 &            20 &           1 &  59.596 & 157.525 & 23.144 &  2.541 \\
 \midrule
2000 &           5 &            20 &          12 &  32.372 &  39.879 & 23.144 &  4.811 \\
2000 &           6 &            20 &          10 &  32.594 &  40.836 & 23.144 &  4.820 \\
2000 &          10 &            20 &           6 &  33.152 &  43.250 & 23.144 &  4.837 \\
2000 &          15 &            20 &           4 &  33.797 &  46.036 & 23.144 &  4.863 \\
2000 &          20 &            20 &           3 &  34.317 &  48.286 & 23.144 &  4.891 \\
2000 &          30 &            20 &           2 &  35.257 &  52.342 & 23.144 &  4.931 \\
2000 &          60 &            20 &           1 &  39.547 &  70.884 & 23.144 &  5.081 \\
 \midrule
5000 &           5 &            20 &          12 &  27.959 &  20.810 & 23.144 & 12.028 \\
5000 &           6 &            20 &          10 &  28.054 &  21.217 & 23.144 & 12.051 \\
5000 &          10 &            20 &           6 &  28.320 &  22.368 & 23.144 & 12.092 \\
5000 &          15 &            20 &           4 &  28.629 &  23.705 & 23.144 & 12.158 \\
5000 &          20 &            20 &           3 &  29.011 &  25.356 & 23.144 & 12.226 \\
5000 &          30 &            20 &           2 &  29.722 &  28.426 & 23.144 & 12.326 \\
5000 &          60 &            20 &           1 &  33.314 &  43.951 & 23.144 & 12.702 \\
\bottomrule
\end{tabular}
\end{sc}
\end{small}
\end{center}
\end{table}
\end{document}